\documentclass[final,onefignum,onetabnum]{siamart190516}

\usepackage{lipsum}
\usepackage{amsfonts}
\usepackage{graphicx}
\usepackage{epstopdf}
\usepackage{algorithmic}

\usepackage{wrapfig}
\usepackage{graphicx}
\usepackage{multirow}
\usepackage{makecell}

\ifpdf
  \DeclareGraphicsExtensions{.eps,.pdf,.png,.jpg}
\else
  \DeclareGraphicsExtensions{.eps}
\fi

\newsiamremark{remark}{Remark}
\newsiamremark{hypothesis}{Hypothesis}
\crefname{hypothesis}{Hypothesis}{Hypotheses}
\newsiamthm{claim}{Claim}

\headers{NPTC-net}{P. Jin, T. Lai, R. Lai and B. Dong}

\title{NPTC-net: Narrow-Band Parallel Transport Convolutional Neural Networks on Point Clouds\thanks{Under review\funding{R. Lai's work is supported in part by an NSF Career Award DMS-1752934. Bin Dong is supported in part by Beijing Natural Science Foundation (Z180001), NSFC 11671022 and Beijing Academy of Artificial Intelligence (BAAI).}}}

\author{
Pengfei Jin\thanks{Beijing International Center for Mathematical Research, Peking University, Beijing, China (\email{jinpf@pku.edu.cn}).}
\and Tianhao Lai\thanks{Beijing International Center for Mathematical Research, Peking University, Beijing, China (\email{howeverlth@pku.edu.cn}).}
\and Rongjie Lai\thanks{Corresponding author. Department of Mathematics, Rensselaer Polytechnic Institute, Troy, NY, USA (\email{lair@rpi.edu}).}
\and Bin Dong\thanks{Corresponding author. Beijing International Center for Mathematical Research, Peking University, Beijing, China; Center for Data Science, Peking University, Beijing, China; and Beijing Institute of Big Data Research, Beijing, China (\email{dongbin@math.pku.edu.cn})}
}

\usepackage{amsopn}

\hypersetup{
  pdftitle={NPTC-net},
  pdfauthor={P. Jin, T. Lai, R. Lai and B. Dong}
}

\begin{document}

\maketitle

\begin{abstract}
  Convolution plays a crucial role in various applications in signal and image processing, analysis and recognition. It is also the main building block of convolution neural networks (CNNs). Designing appropriate convolution neural networks on manifold-structured point clouds can inherit and empower recent advances of CNNs to analyzing and processing point cloud data. However, one of the major challenges is to define a proper way to "sweep" filters through the point cloud as a natural generalization of the planar convolution and to reflect the point cloud's geometry at the same time. In this paper, we consider generalizing convolution by adapting parallel transport on the point cloud. Inspired by a triangulated surface-based method \cite{DBLP:journals/corr/abs-1805-07857}, we propose the Narrow-Band Parallel Transport Convolution (NPTC) using a specifically defined connection on a voxel-based narrow-band approximation of point cloud data. With that, we further propose a deep convolutional neural network based on NPTC (called NPTC-net) for point cloud classification and segmentation. Comprehensive experiments show that the proposed NPTC-net achieves similar or better results than current state-of-the-art methods on point cloud classification and segmentation.
\end{abstract}



\section{Introduction}
\label{sec:introduction}
Data arises from many applications in science and engineering is commonly represented in non-Euclidean structures, such as meshes, point clouds and graphs. Such data includes 3D shapes in computer graphics, scanned point clouds in remote sensing, social networks in social science, functional networks in neuroscience \cite{henaff2015deep,koren2009matrix,Wu_2015_CVPR,morgan1965generation}, etc. These unstructured data often need to be properly analyzed first before they can be effectively utilized in various downstream tasks, such as classification, segmentation, registration, prediction, etc.

Recently, deep learning has enabled major breakthroughs in many fields in science and engineering. This is largely due to the capability of deep neural networks in extracting features, analyzes features and make high level decisions and predictions in an end-to-end fashion. Among different types of deep neural networks, the convolutional neural networks (CNNs) is particularly effective, especially for analyzing Euclidean data such as audios, images, and videos \cite {lecun2015deep, goodfellow2016deep}. Due to the success of CNNs in analyzing Euclidean data, much effort has recently been made to extend CNNs to non-Euclidean data, which is one of the main objectives in geometric deep learning \cite{Masci_2015_ICCV_Workshops, Monti_2017_CVPR, 7974879, cao2020comprehensive}. However, generalizing CNNs to non-Euclidean data is not straightforward. A typical CNN is a composition of simple operators such as convolution operator, down-sampling/pooling operators, batch normalization, etc. Notice that convolution operator, which is the crucial operator in any CNNs, does not admit a simple extension to non-Euclidean data.

This article focuses on generalizing the convolution operator on (manifold structured) point clouds so that it inherits desirable properties of the planar convolution. This in turn enables us to design CNNs on point clouds. In the remaining part of this introduction, we shall first briefly review the applications of point cloud data, and provide a unified view on the existing extensions of convolutions on non-Euclidean data and discuss their relations with our proposed convolution which is called the narrow-band parallel transport convolution.

\subsection{Applications of point clouds data} 

Point cloud data can be acquired by 3D laser scanners, such as Light Detection and Ranging (LIDAR) and RGB-D cameras. It can also be obtained by 3D scene reconstruction from 2D images, such as reconstruction earth's landscape through aerial photography. Furthermore, one can obtain point clouds through sampling of CAD models. Formally, a point cloud $\mathcal{P}=\{x_i\in\mathbb{R}^3: i=1,\ldots,N\}\subset\mathbb{R}^3$ consists of coordinates of points in a $3$-dimensional Euclidean space. Unlike meshes and graphs, point clouds have no connectivity information.

Point clouds provide a direct and convenient way of representing geometric data. For example, in autonomous driving, accurate environment perception is needed to realize reliable navigation and decision in a complex dynamic environment \cite{li2020deep}. Traditionally, image data can provide 2D semantic and texture information with low cost and high efficiency. However, image data lacks 3D geographic information. Therefore, dense and accurate point cloud data with 3D geographic information collected by LIDAR is commonly used by modern autonomous vehicles. In addition, LIDAR is insensitive to the change of lighting conditions and can work both in the daytime nighttime. Other than autonomous driving, many applications prefer to use point cloud data, such as face recognition in e-commerce, urban planning and agricultural production, animation and virtual reality, etc. \cite{liu2019deep}. Due to the vast importance of point cloud data, we would like to extend the convolution operator and design CNNs on point clouds. 

\subsection{Related Work of Convolutions on Non-Euclidean Domains}
Convolution is one of the most widely used operators in applied mathematics, computer science and engineering. It is also the most important building block of Convolutional Neural Netowrks (CNNs) which are the main driven force in the recent success of deep learning. 

In the Euclidean space $\mathbb{R}^n$, the convolution of function $f$ with a kernel (or filter) $k$ is defined as
\begin{equation} \label{convR}
(f \ast k)(x):=\int_{\mathbb{R}^n} k(x-y)f(y) dy.
\end{equation}
This operation can be easily calculated in Euclidean spaces due to the shift-invariance of the space so that the translates of the filter $k$, i.e. $k(x-y)$ is naturally defined. 

One of the main challenges of proposing geometric meaningful convolution on manifolds and point clouds (a discrete form of manifolds) is to define an analogy of the Euclidean translation $x-y$ on the non-Euclidean domain. 
Multiple types of generalized convolutions on manifolds, graphs and point clouds have been proposed in recent years. We shall review some of them and discuss the relation between existing definitions of convolutions and the proposed narrow-band parallel transport convolution.

\textbf{Spectral methods} avoid the direct definition of translation $x-y$ by utilizing the convolution theorem \cite{katznelson2004introduction}: given any two functions $f$ and $g$, $\widehat{f\ast g} = \hat{f} \cdot \hat{g}$. Therefore, we have $f\ast g = (\hat{f} \cdot \hat{g})^{\vee}$, where $\wedge$ and $\vee$ represent generalized Fourier transform and inverse Fourier transform provided through the associated Laplace-Beltrami (LB) eigensystem on manifolds. To avoid computing convolution through full eigenvalue decomposition of the LB operator, polynomial approximation has been proposed and yields convolution as action of polynomials of the LB operator~\cite{HAMMOND2011129,DONG2017452}. Thus, convolutional neural networks can be designed \cite{bruna2014spectral,NIPS2016_6081,8521593}. Spectral methods, however, suffer two major drawbacks. First, these methods define convolution in the frequency domain. As a result, the learned filters are not spatially localized. Second, spectral convolutions are domain-dependent as deformation of the ground manifold will change the corresponding LB eigensystem. This obstructs the use of learning networks from one training domain to a different testing domain~\cite{DBLP:journals/corr/abs-1805-07857}.

\textbf{Spatial mesh-based methods} are more intuitive and similar to the Euclidean case, and this is one of the reasons why most of the existing works fall into this category \cite{Masci_2015_ICCV_Workshops,NIPS2016_6045,DBLP:journals/corr/abs-1805-07857}. The philosophy behind these methods is that the tangent plane $\mathcal{T}_{x}\mathcal{M}$ of a $2$-dimensional manifold $\mathcal{M}$ is embedded to a $2$-dimensional Euclidean domain where convolution can be easily defined. In this paper, we make the first attempt to interpret some of the existing mesh-based methods in a unified framework. We claim that most of the spatial mesh-based methods can be formulated as
\begin{equation} \label{convM}
(f \ast k)(x):= \int_{\mathcal{T}_{x,\epsilon}\mathcal{M}}k(\phi(x,v))f(v){\rm d}v,\quad x\in\mathcal{M}. \end{equation}
Here, $k: \mathbb{R}^2\to \mathbb{R}$ is a convolution kernel and $\mathcal{T}_{x,\epsilon}\mathcal{M} =\{v \in \mathcal{T}_x \mathcal{M}: \left<v,v\right>_{g_x} \le\epsilon^2\}$ with $\epsilon>0$ being the size of the kernel. Given vector fields $\vec{u}^j\in \Gamma(\mathcal{T}\mathcal{M})$, $j=1,2$,  the mapping $\phi(x,\cdot): \mathcal{T}_x\mathcal{M}\to\mathbb{R}^2$ is defined as
\begin{equation}\label{ufunction}
\phi(x,v)  = \left(\left<v,\vec{u}(x)^1\right>_{g_x},\left<v,\vec{u}(x)^2\right>_{g_x}\right),
\end{equation}
Most of the designs of the existing manifold convolutions focused on the designs of $\vec{u}^j$. We remark that possible singularities will lead to no convolution operation at those points. These are isolated points on a closed manifold and do not affect experiment results. In addition, singularities from a given vector field can be overcome using several pairs of vector fields and pooling~\cite{DBLP:journals/corr/abs-1805-07857}.

For example, GCNN \cite{Masci_2015_ICCV_Workshops} and ACNN \cite{NIPS2016_6045} construct a local geodesic polar coordinate system on a manifold, formulating the convolution as $$(f\ast k)(x) = \int k\left((\phi\circ \psi)(\theta,r)\right) (\mathcal{Q}_xf)(r,\theta){\rm d r}{\rm d\theta},$$ where $\mathcal{Q}$ is a local interpolation function with interpolation domain an isotropic disc for GCNN and an anisotropic ellipse for ACNN. A local geodesic polar coordinate system on a manifold can also be transformed to a 2-dimensional planar coordinate system on its tangent plane. Such transformation is the mapping $\psi$ which is defined by the inverse exponential map:
$v = exp^{-1}(z(\theta,r))$ with $z(\theta,r)$ being a point in the local geodesic polar coordinate system at $x\in\mathcal{M}$ with coordinates $(\theta,r)$. With this, we can easily interpret ACNN within the framework of (\ref{convM}). Indeed, ACNN essentially chooses $\vec{u}^j(x)$ as the directions of the principal curvature at point $x$. For GCNN, on the other hand, it avoids choosing a specific vector field on the manifold by taking max-pooling among all possible directions of $\vec{u}(x)^1$ at each point. Such definition of convolution, however, ignores the correspondence of the convolution kernels at different locations.

The newly proposed PTC \cite{DBLP:journals/corr/abs-1805-07857} defines convolution directly on the manifold, while using tangent planes to transport kernels by a properly chosen parallel transport. PTC can be equivalently cast into the form of (\ref{convM}) using the inverse exponential map, and implementation of the proposed parallel transported is realized through choosing specific vector fields $\{\vec{u}^j\}_{j=1,2}$ guided by a Eikonal equation for transforming vectors along geodesic curves on manifolds. Similarly, some definitions of convolution depend on particular frames. PFCNN \cite{yang2020pfcnn} uses the optimized frame as $\{\vec{u}^j\}_{j=1,2}$. The objective function is based on the angles between adjacent frames. CGCNN \cite{yang2021continuous} uses two different data-driven local reference frames as $\{\vec{u}^j\}_{j=1,2}$.

\textbf{Spatial point-based methods} have wider applications due to their weaker assumptions on the data structure.

There are mainly two types of point-based convolution. The first type is to combine the information of points directly. These methods can be formulated as
\begin{equation} \label{convP}
(f \ast k)(x_i):=\sum_{x_j\in \mathcal{N}(x_i)} k(x_i,x_j)f(x_j),
\end{equation}
where $\mathcal{N}(x_i)\subset\mathcal{P}$ is a neighborhood of $x_i$ and kernel $k$ takes different forms in different methods. PointNet \cite{Qi_2017_CVPR} is an early attempt to extract features on point cloud. PointNet is a network structure without convolution, or alternatively we can interpret the convolution defined by PointNet has the simplest kernel $k(x_i,x_j)=\delta(x_i,x_j)$ where $\delta$ is the Kronecker-Delta. Various later works attempt to improve PointNet by choosing different forms of the kernel $k$. For example, PointNet++ \cite{NIPS2017_7095} introduces a max pooling among local points, i.e. choosing kernel $k$ as an indicator function: $k(x_i,x_j) = I_{x_j=\mathop{\arg\max}_{z\in \mathcal{N}(x_i)} f(z)}$. Pointcnn \cite{NIPS2018_7362} chooses $k(x_i,\cdot)=\beta'A_{x_i}$ where $\beta\in\mathbb{R}^K$ and $A_{x_i}\in\mathbb{R}^{K\times K}$ are trainable variables with $K=|\mathcal{N}(x_i)|$. DGCNN \cite{DBLP:journals/corr/abs-1801-07829} proposes an "edge convolution" that can be viewed as fixing $f(x_j)\equiv1$ in (\ref{convP}) and $k(x_i,x_j)=$MLP$(f(x_i),f(x_i)-f(x_j))$, where MLP means the Multi-Layer Perceptron \cite{pal1992multilayer}.

The second type of convolution is defined by first projecting the point cloud locally on an Euclidean domain and then employ regular convolution. This type of methods can also be formulate as (\ref{convM}). For example, Tangent convolutions \cite{Tatarchenko_2018_CVPR} define kernels on the tangent plane, and use 2 principal directions of a local PCA as $\vec{u}_x$. Pointconv \cite{DBLP:journals/corr/abs-1811-07246} constructs local kernels by interpolation in $\mathbb{R}^3$, i.e. letting $\phi(x,v) = x-v $ which is essentially a local Euclidean convolution. Similarly, \cite{wang2018deep,thomas2019kpconv,mao2019interpolated} are also based on local Euclidean convolution. Their main difference are various forms of $k$ and interpolation methods.

\subsection{The Proposed Convolution: NPTC}

We propose a Narrow-Band Parallel Transport Convolution (\textbf{NPTC}) in this paper. It is a geometric convolution based on point cloud discretization of a manifold parallel transport defined in a specific way. As we discussed in the previous section, convolutions in many methods can be written in the form of (\ref{convM}) and (\ref{ufunction}), while the differences mostly lie in the choices of the vector field $\{\vec{u}^j\}_{j}$. As observed by \cite{DBLP:journals/corr/abs-1805-07857} that choosing the vector field properly, the associated convolution can be interpreted as transporting the kernels using the parallel transport associated with the prescribed vector field.

\begin{wrapfigure}{r}{0.5\textwidth}
  \centering
  \includegraphics[width=.8\linewidth]{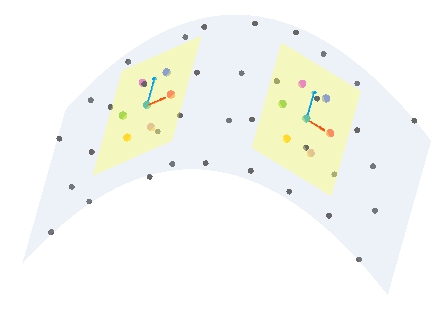}
  \caption{Narrow-band parallel transport convolution on point cloud: black and blue points are sample points of the surface (blue). Each kernel (colored dots) is defined on the tangent plane (yellow). The vectors on the tangent planes that are of the same color are defined by parallel transport.}
\end{wrapfigure}

We attempt to define geometric convolutions that can be viewed as translating kernels on the point cloud in a parallel fashion. One naive approach to extend mesh based methods to point cloud is to generate a triangulated surface based on the point cloud. However, this is not as convenient as working directly with the point cloud since in practice not every point cloud corresponds to a legitimate parameterized surface and  pooling is not as easy to implement on triangulated surfaces as on point clouds. In addition, it is time-consuming to construct meshes on point clouds. When applied in practice, mesh construction time may be much longer than inference time of some methods directly applied on the point cloud. To avoid mesh constuction and to handle point clouds data directly, we propose to define convolution on point clouds by combining voxelization and geometric convolution.

Now, we describe how NPTC is computed on point clouds. Firstly, point clouds are approximated by voxel-based narrow-band with appropriate resolution. For a manifold $\mathcal{M}$ or point cloud $\mathcal{P}$, its narrow-band is defined as the region with distance from $\mathcal{M}$  (or $\mathcal{P}$) less than $\epsilon$: $\mathcal{NB}(\mathcal{M}):=\{x\in \mathbb{R}^3|dist(x,\mathcal{M})<\epsilon\}$, where $dist$ represents a distance function. For calculation efficiency and robustness to noise, we choose voxel-based narrow-band approximation. It should be noted that the result of the traditional voxelization method \cite{Wu_2015_CVPR} is a solid 3-d structure and the voxel-based narrow-band can be understood as the "shell" with a certain thickness of the former (if the point cloud is sampled from the surface of the object). It is very efficient to calculate the distance function in voxel approximation with appropriate resolution. The vector field $\{\vec{u}_x\}_{x\in \mathcal{M}}$ is defined as the projections of the gradient field of a narrow-band-based distance function on the approximated tangent plane of the point cloud. Such definition of the vector field is robust to noise, since if the distortion of the coordinates of the points by noise does not exceed the width of the narrow-band, the computed gradient field of the distance function in the narrow-band remains unchanged. After the vector field $\{\vec{u}_x\}_{x\in \mathcal{M}}$, we use local PCA to estimate normal vectors on the point cloud following \cite{Xu_2018_ECCV}. Finally, the convolution kernel can be constructed on the tangent plane and the corresponding $f(v)$ can be obtained by interpolating the value $f(x)$ on the point cloud.

Note that we prefer to use geometric convolution in NPTC because compared with methods that translate kernels in the ambient space of the manifold, NPTC translates kernels on the tangent planes which effectively avoids having convolution kernels defined away from the underlying manifold of the point cloud. In other words, NPTC can well reflect point cloud geometry and is a natural generalization of planar convolution in the sense that when the point cloud reduces to planar grids, the NPTC reduces to the planar convolution.

\subsection{Contributions}

\begin{itemize}
\item We introduce a new point cloud convolution, NPTC, based on parallel transport defined by a narrow-band approximation of the point cloud. The proposed NPTC is a natural generalization of planar convolution.

\item The proposed NPTC combines voxelization and geometric convolution. Voxelization with appropriate resolution brings robustness and geometric convolution can better reflect point cloud's geometry.

\item Based on NPTC, we designed convolutional neural networks, called NPTC-net, for point clouds classification and segmentation with state-of-the-art performance.
\end{itemize}

The rest of this paper is organized as follows. In section \ref{sec:background}, we will discuss the mathematical background of parallel transports on manifolds and the Eikonal equation for computing distance functions. In section \ref{sec:NPTC}, we propose the narrow-banded parallel transport convolution and it is associated with convolutional neural networks on point cloud represented manifolds. After that, we report our intensive numerical experiments of point clouds classification and segmentation on benchmark data sets in section \ref{sec:experiment}. We conclude the paper in \ref{sec:conclusion}.

\section{Background}
\label{sec:background}
\subsection{Manifold and Parallel Transport}

Let $\mathcal{M}$ be a two-dimensional differential manifold embedded in $\mathbb{R}^3$. We write  $\mathcal{T}_{x} \mathcal{M}$ as the two-dimensional tangent plane at point $x\in\mathcal{M}$. The disjoint union of the tangent planes at each point on the manifold defines the tangent bundle $\mathcal{T}\mathcal{M}$. A vector field $V$ is a smooth assignment: $\mathcal{M} \rightarrow \mathcal{T}\mathcal{M}$ such that $V(x)\in \mathcal{T}_x\mathcal{M}, \forall x \in \mathcal{M}$. The collection of all smooth vector fields is denoted as $\Gamma( \mathcal{T} \mathcal{M})$.

An affine connection is a bilinear mapping $\nabla$: $\Gamma( \mathcal{T} \mathcal{M}) \times \Gamma( \mathcal{T} \mathcal{M}) \rightarrow \Gamma( \mathcal{T} \mathcal{M})$, such that for all smooth functions $f$ and $g$ in $C^{\infty}(\mathcal{M})$ and all vector fields $U,V$ and $W$ on $M$:
\begin{equation}
\left\{
\begin{array}{l}
\nabla_{fU + gV} W = f \nabla_U W + g \nabla_V W, \\
\nabla_{U} (aV + bW) = a \nabla_U V + b \nabla_U W, \quad a, b \in \mathbb{R}, \\
\nabla_U (fV) = df(U)V+f\nabla_U V. \\
\end{array}
\right.
\end{equation}
A vector field $U$ is called \textbf{parallel} along a curve $\gamma:I\rightarrow \mathcal{M}$ if $\nabla_{\dot \gamma} U = 0$. Given an vector $\vec{e} \in \mathcal{T}_{x_0}\mathcal{M}$ at $x_0 = \gamma(0) \in \mathcal{M}$, the \textbf{parallel transport} of $\vec{e}$ along $\gamma$ is the extension of $\vec{e}$ to a parallel section $U$ on $\gamma$. More precisely, $U$ is the unique section of $\Gamma( \mathcal{T} \mathcal{M})$ along $\gamma$ satisfying the ordinary differential equation $\nabla _{\dot{\gamma}(t)}U(t) = 0$ with the initial value $U(0) = \vec{e}$.

In differential geometry, a geodesic on a smooth manifold $\mathcal{M}$ with an affine connection $\nabla$ is a curve $\gamma(t)$ such that parallel transport along the curve preserves the tangent vector to the curve. It is a generalization of the notion of a "straight line". Formally, a geodesic is $\gamma:[0,l]\rightarrow\mathcal{M}$ if $\nabla _{\dot{\gamma}(t)}\dot{\gamma}(t) = 0$. For any two points $x_0$ and $x_1$ on a closed manifold $\mathcal{M}$, there will be a geodesic connecting $x_0$ and $x_1$. More details on aforementioned concepts can be referred in \cite{kobayashi1969foundations}.

Alternatively, an affine connection can be defined by an assignment $\Xi$ as a family of linear transformations on tangent spaces along any smooth curve on $\mathcal{M}$. Consider $\gamma^y_x$ a smooth arc joining two, not necessarily distinct, points from $x$ to $y$, define $\Xi(\gamma^y_x):\mathcal{T}_x\mathcal{M} \rightarrow \mathcal{T}_y\mathcal{M}$.  If $\Xi(\gamma^y_x)$ satisfies the following properties:
\begin{itemize}
  \item [1)] 
  $\Xi(\gamma^y_x)$ is non-singular,       
  \item [2)]
  $\lim_{y\rightarrow x}\Xi(\gamma^y_x)=Id$,
  \item [3)]
  $\Xi(\gamma^z_x)=\Xi(\gamma^z_y)\Xi(\gamma^y_x)$,
  \item [4)]
  $\Xi$ is Frech\'{e}t differentiable in terms of $\gamma$, $x$ and $y$. 
\end{itemize}
Then, the vector $\vec{v}_y$ is obtained by parallel displacement of $\vec{v}_x$ along $\gamma^y_x$ is provided as $\vec{v}_y=\Xi(\gamma^y_x)\vec{v}_x$. Condier $V$ is a tangent vector field along $\gamma$, the associated infinitesimal connection $\nabla_{\dot{\gamma}}V = \lim_{h\rightarrow 0}\frac{1}{h} (\Xi(\gamma^{\gamma(0)}_{\gamma(h)})V_{\gamma(h)} - V_{\gamma(0)})$ can be induced from $\Xi$ \cite{knebelman1951spaces}.

For convenience, instead of transporting the kernel on the manifold, we can locally construct the parallel transported kernel at every point $x$ by formulating $k$ as $k(\phi(x,\cdot))$, where $\phi(x,v)$ is defined in (\ref{ufunction}). It is known in differential geometry that transporting the kernel to every point on the manifold is the same as locally reconstructing the kernel in the aforementioned way.

\subsection{Solving Eikonal Eqution on points clouds}

As mentioned before, the proposed NPTC relies on the computation of a distance function on a voxel-based narrow-band approximation of the given point cloud. Therefore, we briefly review what a narrow-band approximation is and how the distance function is calculated.

Geometric attributes calculation of large unstructured point-based data sets is a challenging task. Hence, the most common way of
dealing with unstructured point-based data is to re-sample to a structured grid \cite{franke1991scattered}. A large variety of well-known methods like isosurface extraction \cite{lorensen1987marching},
region-growing methods \cite{hojjatoleslami1998region}, and level-set methods \cite{osher1988fronts,osher2006level} can be applied to gridded data. The original idea of level sets is to implicitly represent a surface as the solution of an equation concerning an underlying scalar field. To improve computation efficiency of level-set methods, local level-set method was introduced \cite{adalsteinsson1995fast} which essentially suggests conduct computations only within a narrow-band of the zero level set.

Different methods are proposed to obtain narrow-band representation of shapes in 2D \cite{jiang2012improve, bindu2014fast} or 3D \cite{rosenthal2008smooth, rosenthal2011narrow}. In this work, the narrow-band of a point cloud $\mathcal{P}$ is defined as the region with distance from $\mathcal{P}$ less than $\epsilon$: $\mathcal{NB}(\mathcal{P}):=\{x\in \mathbb{R}^3|dist(x,\mathcal{P})<\epsilon\}$, where $dist$ represents the standard Euclidean distance function.

We follow the previous work and consider the gridded narrow-band. Consider the point cloud with $N$ points in the unit cube and divide the unit cube into $M^3$ small cubes. The side length of each small cube is $\frac{1}{m}$. The set of all small cubes whose distance from the point cloud is less than $\epsilon$ form the voxel-based narrow-band approximation of the point cloud. The number of grid points in traditional voxelization is $M^3$, and the number of grid points in the voxel-based narrow-band for a smooth surface is $O(kM^2)$, and $k$ depends on the thickness of the narrow-band.

Distance functions can be easily computed by solving the Eikonal equation \cite{eldar1997farthest}. The Eikonal equation is a non-linear partial differential equation describing wave propagation:
\begin{equation} \label{Eikonal}
|\nabla \rho| = 1/h(x), \quad x\in\Omega,\quad \rho|_{\Lambda} = 0,
\end{equation}
where $\Lambda\subset\overline{\Omega}\subset\mathbb{R}^n$ and $h(x)$ is a strictly positive function. The solution $\rho(x)$ of (\ref{Eikonal}) can be viewed as the shortest time needed to travel from $x$ to $\Lambda$ with $h(x)$ being the speed of the wave at $x$. For the special case when $h=1$, the solution $\rho(x)$ represents the distance from $x$ to $\Lambda$ limited in the $\Omega$. The Eikonal equation can be solved by the fast marching method \cite{sethian1996fast} or the fast sweeping method \cite{zhao2005fast}. Both methods have an optimal computation complexity of $O(L)$ with $L$ being the number of the grid points. Naturally, when $N\gg M^2$, solving the equation in the voxel-based narrow-band will be faster than calculating directly on the point cloud or the classical voxelization. In fact, in practical applications, such as LiDAR data, the number of points in a point cloud is often far greater than the resolution required to adequately represent the shape of interest.

\section{Narrow-band Parallel Transport Convolution (NPTC) and Network Design}
\label{sec:NPTC}
Generalization of convolution defined by parallel transport on triangulated surfaces has already been proposed in \cite{DBLP:journals/corr/abs-1805-07857}. In this section, we discuss how to transport kernels on point clouds in a similar fashion.

\subsection{Narrow-band Parallel Transport Convolution (NPTC)}

For a given function $f:\mathcal{P}\to\mathbb{R}$, the NPTC of $f$ with kernel $k$ takes the same form as (\ref{convM}). Under such formulation, the key to design a convolution is to design vector fields $\{\vec{u}^j\}_{j=1,2}$. In this subsection, we discuss the general idea of NPTC and the interpretation of it in terms of parallel transport.

\subsubsection{General Idea of NPTC}

To select a suitable vector field, we first recall the choice of the vector fields of PTC which defines convolution on triangulated surfaces via parallel transport with respect to the Levi-Civita connection \cite{DBLP:journals/corr/abs-1805-07857}. Geodesic curve represents the shortest path between two points on a Riemannian manifold. Given a geodesic connecting two points $x$ and $y$, the tangential direction at $x$ corresponds to the ascent direction of geodesic distance from $y$. PTC chooses such direction as $\vec{u}_x^1$ and defines $\vec{u}_x^2=\vec{u}_x^1\times \vec{n}_x$ where $\vec{n}_x$ is the normal vector at $x$.

To construct a vector field on a point cloud, we also consider to use gradient of a distance function on the point cloud. However, unlike triangulated surfaces, distance function is not easily defined on point clouds due to the lack of connectivity. It is then natural to approximate the point cloud with another data structure with connectivity, so that distance function can be easily calculated. For convenience and efficiency, we use voxelization \cite{Wu_2015_CVPR} to approximate the point cloud in a narrow-band in $\mathbb{R}^3$ covering the point cloud. We denote such distance function as $\rho:\mathbb{R}^3\to\mathbb{R}^+$. We will elaborate how $\rho$ can be calculated in later parts of this subsection.

Note that if the point cloud is sampled from a plane, the narrow-band is flat as well. Then, by a proper choice of the distance function, the vector fields $\{\vec{u}^j\}_{j=1,2}$, can be reduced to the global coordinate $\{\vec{e}_j\}, j=1,2$ on the plane. This means that NPTC is reduced to the traditional planar convolution.

Once the distance function $\rho$ is computed, we choose $\vec{u}_x^1=\nabla_{\mathcal{P}}\rho(x)$, where $\nabla_{\mathcal{P}}\rho(x)$ is a projection of $\nabla \rho(x)$ on an approximated tangent plane at $x$. Then, $\vec{u}_x^2$ can be calculated by the outer product $\vec{u}_x^2=\vec{u}_x^1\times \vec{n}_x$ with $\vec{n}_x$ the normal vector at $x$. The value $f(v)$ is computed by nearest-neighbor interpolation \cite{grevera1998objective}, i.e., $f(v)=f(z)$ where $z\in\mathcal{P}$ is the closest point to $v$. Note that one may use a more sophisticated method to compute $f(v)$ rather than using nearest-neighbor interpolation. We choose nearest-neighbor interpolation because of its simplicity.

\begin{figure}[h!]
\begin{center}
\includegraphics[width=360pt]{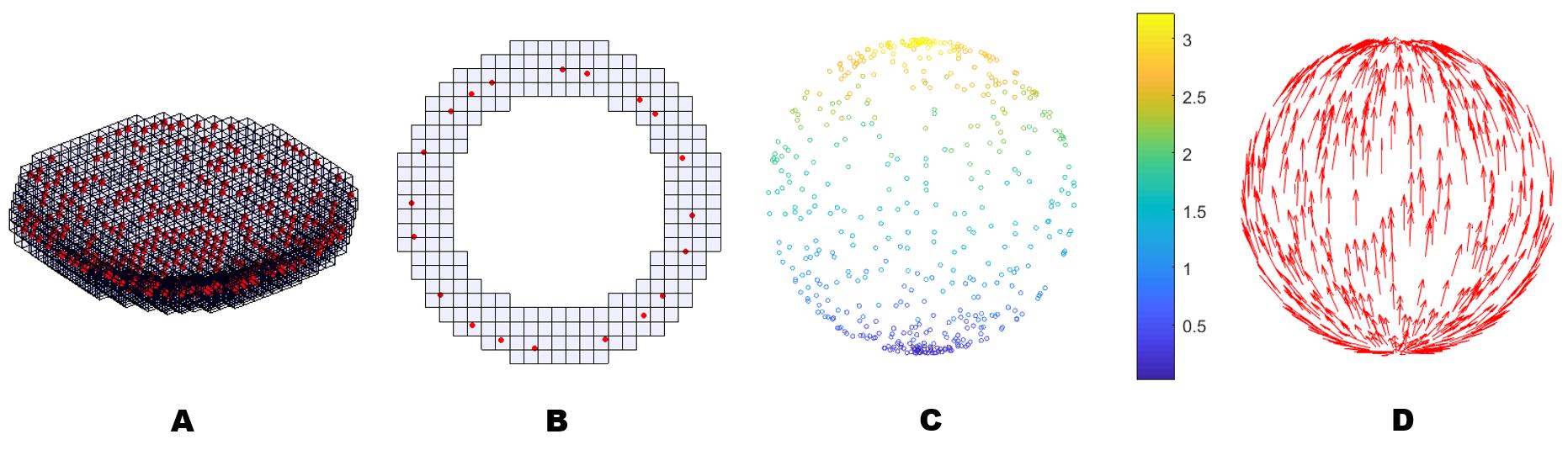}
\end{center}
\caption{Illustration of a point cloud $\mathcal{P}$ sampled from the unit sphere. (A) shows the narrow-band approximation (blue boxes) of part of $\mathcal{P}$ (in red). (B) is a cross section of (A). (C), (D) show the distance function $\rho$ and vector field $\{\vec{u}_x^1\}$ ($\{ \nabla_{\mathcal{P}}\rho(x) \}$) on the point cloud. We can see that distance propagates from the bottom center to the top center reflecting the geometry of the sphere.}\label{demo}
\end{figure}

\subsubsection{Computing Distance Function on Point Clouds}

A point cloud is entirely discrete without inherent connectivity. Therefore, it is not straightforward to compute the distance function on point clouds although the local mesh method~\cite{lai2013local} can be applied to solve the Eikonal equation. For simplicity, we use voxels to approximate point clouds and to compute distance functions on the voxels using the well-known fast marching method based on regular grid provided by the voxelization~\cite{sethian1996fast}. Note that, using voxels to compute distance functions is fast and robust to noise and local deformations.

The solution $\rho(x)$ of the Eikonal equation $|\nabla \rho(x)| = 1$ presents the distance form $\Lambda$ to $x$ limited inside the narrow-band. Here $\Lambda$ is chosen as a certain point on the point cloud. We note that the starting point will become a singularity, but on non-parallelizable manifolds such as the sphere, it is not possible to construct a smooth vector field \cite{helgason1979differential}. So that the vector field must have at least one a singularity. Although generating multiple vector fields by selecting different starting points is helpful to eliminate singularities, experiments show that directly selecting one point already provides satisfactory results. Also, the ensemble result of several vector fields has almost no improvement compared with each of a single vector field.

Finally, we interpolate the distance function from the voxels to the point cloud.

\subsubsection{Computing the Vector Fields on Point Clouds}

We first compute the tangent plane on each point. Tangent planes are important features of manifolds and have been well-studied in the literature \cite{lai2013local}. In this paper, we use local principal component analysis (LPCA) to estimate the tangent plane. We estimate the local linear structure near $x\in\mathcal{P}$ using the covariance matrix $$\sum_{x_k \in \mathcal{N}(x)} (x_k-c)^\top (x_k-c),\quad c=\frac{1}{k}\sum_{x_k \in \mathcal{N}(x)} x_k,$$ where $\mathcal{N}(x)$ is the set of neighboring points of $x$. The eigenvectors of the covariance matrix form an orthogonal basis. If the point cloud is sampled from a two dimensional manifold, and the local sampling is dense enough to resolve local features, the eigenvectors corresponding to the largest two eigenvalues provide the two orthogonal directions of the tangent plane, and the remaining vector represents the normal direction at $x\in\mathcal{P}$. Here, we denote the space spanned by the two eigenvectors of the covariance matrix at $x$ as $\mathcal{T}_{x}\mathcal{P}\subset\mathbb{R}^3$.

With the computed distance function $\rho(x)$, it is nature to define the vector field by projecting $\nabla  \rho$ on the approximated tangent planes of the point cloud. Given a point $x_k\in\mathcal{P}$ close enough to $x\in\mathcal{P}$, we have $$\left< \nabla \rho , x_k-x \right> \approx \rho(x_k) - \rho(x),$$ where $\rho(x_k)$ and $\rho(x)$ are known. If we consider $k$-nearest neighbors of $x$, we have $k-1$ equations with 3 unknowns that are the three components of $\nabla \rho$. We can use least squares to find $\nabla \rho$. We then project the vector $\nabla \rho(x)$ onto the tangent plane at $x$. We denote the projected vector $\nabla_{\mathcal{P}}\rho$, which is the vector we eventually need to define NPTC as described in Section 3.1.1. Denote the calculated vector field as $\{\vec{u}^j\}_{j=1,2}$, and the starting point of the distance function as $x_0$. For any point $x$, let $\gamma_{x_0}^x$ be the integral curve of $\nabla\rho$  connecting $x$ and $x_0$. Define a linear transformation $\Xi(\gamma_{x_0}^x)$ satisfying $\vec{u}^j_x=\Xi(C_{x_0}^x)\vec{u}^j_{x_0},j=1,2$. As mentioned in the section \ref{sec:background}, defining linear transformations on any arc satisfying specific properties means defining connection and parallel transport. In applications, we do not need linear transformations along all arcs. NPTC is defining linear transformations on integral curves of $\nabla\rho$ (as approximated geodesic curves). 

However, the quality of vector fields is affected by the quality of the estimated normal directions. On natural concerns is how such estimation is affected by noise. We remark that LPCA is in fact quite robust to noise. In the section of experiments, we will observe that our method can still achieve satisfactory results on the real-world scanned dataset with noise.

\subsection{NPTC-net: Architecture Design for Classification and Segmentation}

This section, we present how to use NPTC to design convolutional neural networks on point clouds for classification and segmentation tasks. For that, other than the NPTC, we need to define some other operations that are frequently used in neural networks. 

\begin{figure}[h!]
  \centering

  \includegraphics[width=.8\linewidth]{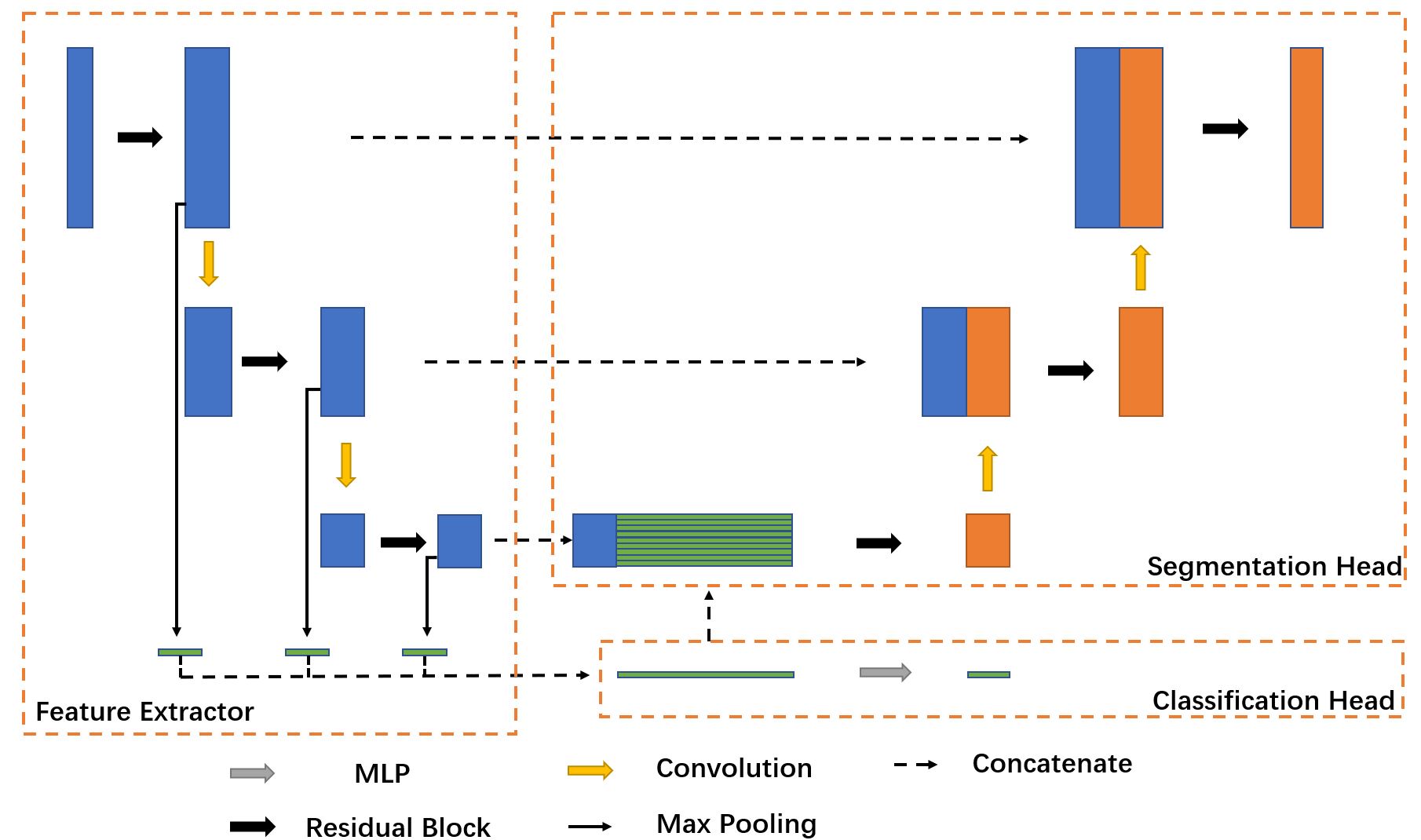}
  \caption{Architecture of NPTC-net. The network for segmentation tasks is at the top right with
encoding and the network for classification tasks is at the bottom right with encoding.}
  \label{arch}
\end{figure}

\textbf{Down-sampling:} Given the input point cloud $\mathcal{P}$, we use farthest point sampling \cite{eldar1997farthest} to generate a sequence $\{\mathcal{P}^{(i)}\}$, where $\mathcal{P}^{(0)}=\mathcal{P}$ and $\mathcal{P}^{(i)} \subset \mathcal{P}^{(i-1)}$ represents the $i$-th hierarchical structure of $\mathcal{P}$. The down-sampled set of $\mathcal{P}^{(i)}$ is $\mathcal{P}^{(i+1)}$ and up-sampled set is $\mathcal{P}^{(i-1)}$ during convolution layer. In $k$-th layer, feature maps is denoted as $\mathcal{F}^{k}\in \mathbb{R}^{N^k\times c^k}$, where $N^k$ is the number of the points and $c^k$ is the number of channels at layer $k$. 

\textbf{Multi-Layer Perceptron(MLP):} Given the feature maps $\mathcal{F}$, $\text{MLP}(\mathcal{F})$ is defined as $\sigma(...\sigma(\mathcal{F}W^1)W^2...)W^L)$, where $\sigma$ is a non-linear function and $W^i$ is a matrix.

\textbf{Concatenate:} Given the feature maps $\mathcal{F}_1\in \mathbb{R}^{N\times c_1}$ and $\mathcal{F}_2\in \mathbb{R}^{N\times c_1}$, concatenate of $\mathcal{F}_1$ and $\mathcal{F}_2$ is defined as $\mathcal{F}_3=[\mathcal{F}_1;\mathcal{F}_2]\in\mathbb{R}^{N\times (c_1+c_2)}$.

\textbf{Global max pooling:} Given the feature maps $\mathcal{F}\in \mathbb{R}^{N\times c}$, $\hat{f} \in \mathbb{R}^c$ is the global max pooling of $\mathcal{F}$ satisfying $\hat{f}_j=\max\limits_{i}{\mathcal{F}_{ij}}$.

\textbf{Convolution layer:} Our $k$-th convolution layer takes points $\mathcal{P}^k \in \mathbb{R}^{N^k\times 3}$ and their corresponding feature maps $\mathcal{F}^{k}\in \mathbb{R}^{N^k\times c^k}$ as input. The corresponding output is $\mathcal{F}^{k+1}\in \mathbb{R}^{N^{k+1}\times c^{k+1}}$ living on the points $\mathcal{P}^{k+1} \in \mathbb{R}^{N^{k+1}\times 3}$. The NPTC-net have encoding and decoding stages. Normally, $N^{k+1} \leq N^k$ during encoding and $N^k \geq N^{k+1}$ during decoding. Convolution at the $k$-th layer during encoding is only performed on the point set $\mathcal{P}^{k+1}$, which resembles convolution with stride $\geq 1$ for planar convolutions.

\textbf{Residual block:} One residual block takes the feature maps $\mathcal{F}\in \mathbb{R}^{N\times c}$ on the point set $\mathcal{P} \in \mathbb{R}^{N\times 3}$ as input and same number of points and same number of channels of features as output. One residual layer consists of three components: MLP from $c$ channels to $\frac{c}{2}$ channels, convolution layer from $\frac{c}{2}$ channels to $\frac{c}{2}$ channels, MLP from $\frac{c}{2}$ channels to $c$ channels plus the feature maps from the bypass connection. A residual block consists of several residual layers.

\textbf{NPTC-net} consists of the aforementioned operations and its architecture is given by Figure \ref{arch}. The left half of the NPTC-net is the encoder part of the network for feature extraction. For classification, features at the bottom of the network are directly attached to a classification network; while for segmentation, features are decoded using the right half of the NPTC-net (decoder part of the network) to output the segmentation map. Each rectangle represents a feature map, the height of rectangle represents the number of points, and the width represents the number of channels.

\section{Experiments}
\label{sec:experiment}
In order to evaluate our new NPTC-nets, we conduct experiments on three widely used 3D datasets, ModelNet \cite{Wu_2015_CVPR}, ShapeNet Part \cite{Yi:2016:SAF:2980179.2980238}, S3DIS \cite{7780539}. ModelNet40 and ShapeNet Part contain point clouds generated from CAD models, S3DIS is a real-world scanned dataset. ModelNet40 is for 3D shape classification, and ShapeNet Part and S3DIS are for segmentation (or partition) of 3D shapes.


\subsection{Implementation Details}

We implement the model with Tensorflow\cite{abadi2016tensorflow}. The neural network is denoted as $f_\theta$, where $\theta$ corresponds to the trainable parameters of the network. Given labeled data pair $\{P_i,y_i\}$, where $P_i$ corresponds to the point cloud data (i.e. the collection of 3D points of a given geometric object) and $y_i$ is its associated label, the learning tasks is formulated as the optimization problem $\min\limits_{\theta} \sum_i L(f_\theta(P_i),y_i)$ for some loss function $L$. For classification, $L$ is the cross-entropy loss: $L(f_i, y_i)=-\sum_{i=1}^{c} y_{ij}log(f_{ij})$, where $y_{ij}$ corresponds to the $j$-th element of one-hot encoded label of $y_i$ and $f_{ij}$ denotes the $j$-th element of $f_i$. Segmentation task can be viewed as point-wise classification task with cross-entropy loss.

We using SGD optimizer \cite{robbins1951stochastic} with an initial learning rate $0.1$ for ModelNet40 and ADAM optimizer \cite{DBLP:journals/corr/KingmaB14} with an initial learning rate $0.002$ for ShapeNet Part and S3DIS on a GTX TITAN Xp GPU. 

To avoid over-fitting, data augmentation \cite{shorten2019survey} is used during training. Data augmentation is to generate more data pairs $\{\mathcal{A}_j(P_i),y_i\}$, where $\mathcal{A}_j$ corresponds to some transformation without changing the label. For the classification task, the data is augmented by random rotation, scaling and Gaussian perturbation on the coordinates of the points. For segmentation, the data is augmented by scaling and Gaussian perturbation on the coordinates of the points.

We do inferences on the augmented data during testing and aggregate the results by voting following \cite{NIPS2017_7095}. Let $\hat{y}_{ij} = f_\theta(\mathcal{A}_j(P_i))$ denote the probability distribution of the predict on $\mathcal{A}_j(P_i)$. The aggregated inference result of $P_i$ is $\hat{y}_i = \frac{1}{n_a}\sum_{j=1}^{n_a} \hat{y}_{ij}$.

Following \cite{NIPS2018_7362}, the data of S3DIS is firstly split by room, and then the rooms are sliced into $1.5m$ by $1.5m$ blocks, with $0.3m$ padding on each side. The points in the padding areas serve as context of the internal points, and themselves are not linked to loss in the training phase, nor used for prediction in the testing phase. Each block will be viewed as a point cloud during training and testing.

\subsection{3D Shape Classification and Segmentation}

We test the NPTC-net on ModelNet40 for classification tasks. ModelNet40 contains 12,311 CAD models from 40 categories with 9,842 samples for training and 2,468 samples for testing. For comparison, we use the data provided by \cite{Qi_2017_CVPR} sampling 2,048 points uniformly and computing the normal vectors from the mesh. As shown on Tabel \ref{all}, our networks outperform other state-of-art methods. (If a compared method has results on both 2048 (or 1024) and 5000 points, we only compare with the former.).

We also evaluate the NPTC-net on ShapeNet Part for segmentation tasks. It contains 16,680 models from 16 shape categories with 14,006 for training and 2,874 for testing, each annotated with 2 to 6 parts and there are 50 different parts in total. We follow the experiment setup of previous works, putting object category into networks as known information. We use point intersection-over-union (IoU) to evaluate our NPTC-net. Table \ref{all} shows that our model ranks second on this dataset and is fairly close to the best known result.

\begin{table}[h]
  \caption{Comparisons of overall accuracy (OA) and mean per-class accuracy (mA) on ModelNet40 as well as comparisons in instance average IoU (mIoU) and class average IoU (mcIoU) on ShapeNet Part. Models ranking first is colored in red and second in blue.}
  \label{all}
  \centering
  \begin{tabular}{ccccc}
\hline
   &\multicolumn{2}{c}{Modelnet40} &\multicolumn{2}{c}{ShapeNet part}\\
\hline

    Method  & OA(\%)    &mA(\%)    &  mIoU     & mcIoU  \\
    \hline
kd-net \cite{8237361} & 91.8& \textcolor{blue}{88.5} & 82.3 & 77.4 \\


pointnet \cite{Qi_2017_CVPR} & 89.2& 86.2& 83.7 & 80.4\\
SO-Net \cite{Li_2018_CVPR} & 90.9& 87.3& 84.9 & 81.0\\
pointnet++ \cite{NIPS2017_7095} & 90.7&  -& 85.1 &81.9\\

SpecGCN \cite{Wang_2018_ECCV} & 92.1 &- & 85.4& -\\
SpiderCNN \cite{Xu_2018_ECCV} & \textcolor{blue}{92.4}& -& 85.3 & 81.7\\
pointcnn \cite{NIPS2018_7362} & 92.2& 88.1&\textcolor{red}{ 86.1 }& \textcolor{red}{84.6}\\
Ours & \textcolor{red}{92.7}  & \textcolor{red}{ 90.2} & \textcolor{blue}{85.8} & \textcolor{blue}{83.3} \\
    \hline
  \end{tabular}
\end{table}

\subsection{3D scene semantics segmentation}
As pointed out earlier, using voxelization can bring rubustness to the definition of geometric convolution.
In order to show the robustness of our model for real data, we tested scene semantics segmentation on the "Stanford Large-Scale 3D Indoor Spaces Dataset" (S3DIS). S3DIS covers six large-scale indoor areas from 3 different buildings for a total of 273 million points annotated with 13 classes. This is a real-world scanned dataset without normal and with noise. Following \cite{8374608}, we advocate the use of Area-5 as test scene to better measure the generalization ability of our method. Table \ref{S3DIS} shows that geometric convolution methods are close to the methods which do not need normal estimation. NPTC-net outperforms the best known geometric convolution method TangentConv\cite{Xu_2018_ECCV}.

\begin{table}[h]
  \caption{Comparisons of overall accuracy (OA) and mean per-class IoU (mIoU) on S3DIS. Models ranking first is colored in red and second in blue.}
  \label{S3DIS}
  \centering
  \begin{tabular}{cccc}
    \hline
Convolution Type & Method   &  OA(\%)  &   mIoU(\%) \\
   \hline
no convolution & pointnet \cite{Qi_2017_CVPR} &78.8 & 41.3 \\
   \hline
\multirow{3}{1.2in}{\ $3$-d convolution} & SegCloud \cite{8374608}&- & 48.9 \\
 & Eff3DConv\cite{zhang2018efficient} & 69.3 & 51.8 \\

 &ParamConv\cite{wang2018deep} & - & \textcolor{red}{58.3} \\
 \hline
\multirow{2}{1.6in}{geometric convolution}  & TangentConv \cite{Xu_2018_ECCV}&\textcolor{blue}{82.5} & 52.8  \\
 & Ours& \textcolor{red} {83.7} & \textcolor{blue}{54.0}\\
\hline
  \end{tabular}
\end{table}

\subsection{Visualization of Segmentation}
\begin{figure}[h!]
  \centering

  \includegraphics[width=360pt]{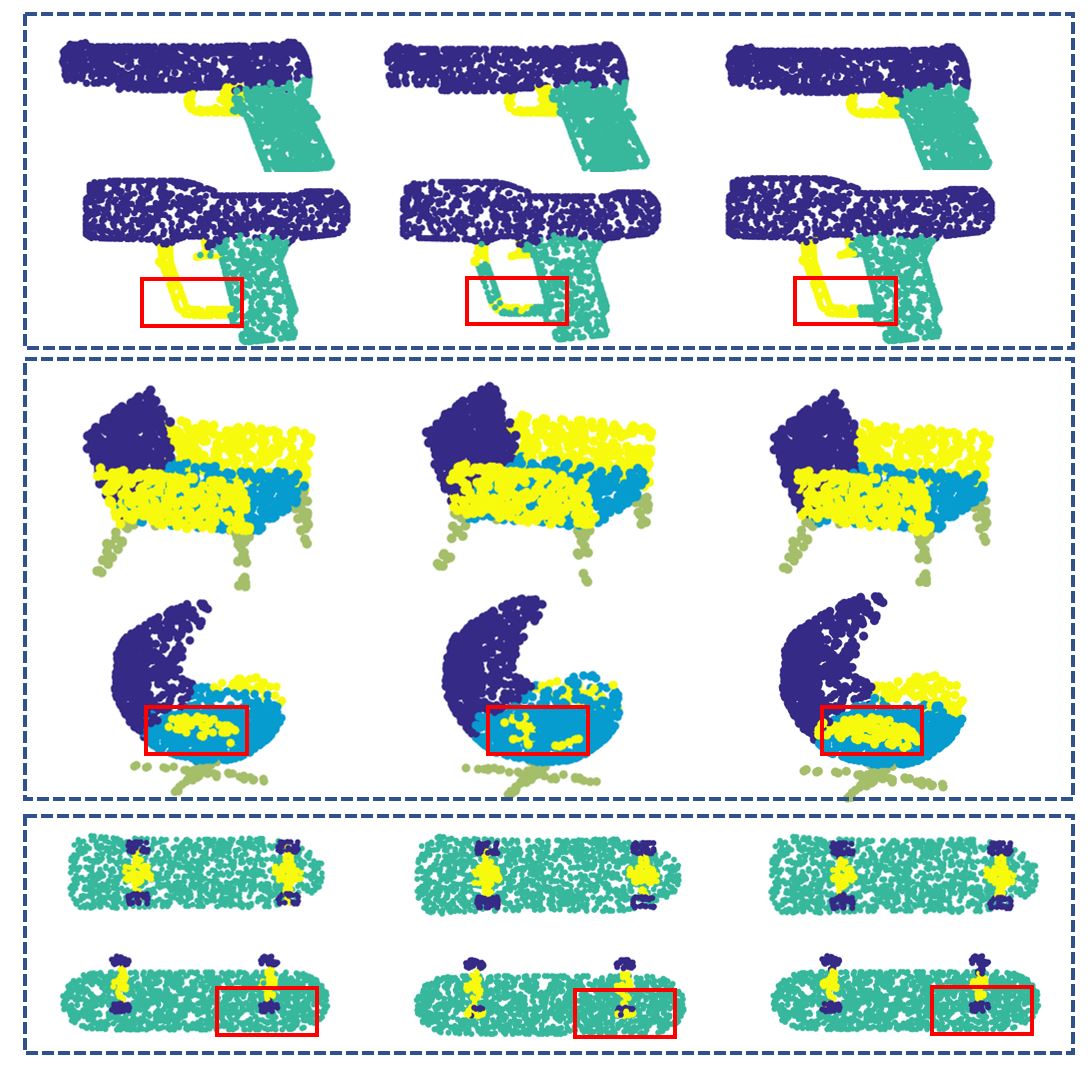}
  \caption{Comparison of segmentation results of unusual structures: three columns from left to right represent the label, the prediction of pointnet++, and the prediction of NPTC-net.}
\label{seg_out}
\end{figure}

\begin{figure}[h!]
  \centering

  \includegraphics[width=360pt]{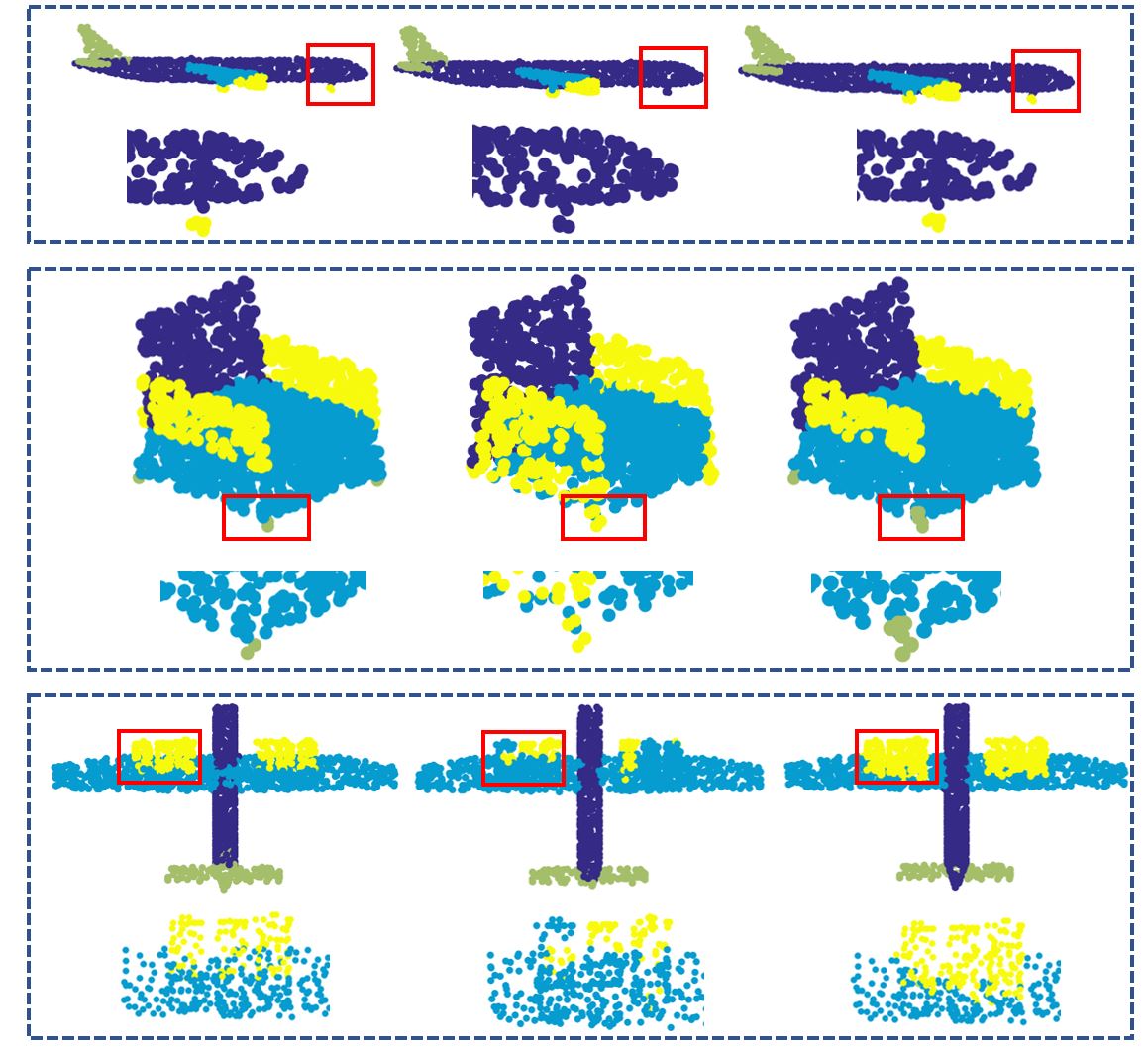}
  \caption{Comparison of segmentation results of small structures: three columns from left to right represent the label, the prediction of pointnet++, and the prediction of NPTC-net.}
\label{seg_tiny}
\end{figure}

To better demonstrate the advantages of our model, we visualize the segmentation results of the test data in the ShapeNet Part. We compare NPTC-net with the popular model Pointnet++ \cite{NIPS2017_7095} and select several representative examples. In Figures \ref{seg_out} and \ref{seg_tiny}, three columns from left to right represent the ground truth, the prediction by pointnet++, and the prediction by NPTC-net. Each blue box of Figure \ref{seg_out} contains objects within the same class (handgun, chair and skateboard). In each blue box, the first row contains objects with standard structures while the second row contains objects with certain unusual structures which is highlighted by red boxes. In each blue box of Figure \ref{seg_tiny}, the first row contains the overviews of the objects while second row contains the corresponding zoom-in views.

As shown in Figure \ref{seg_out}, we found that both models can make good predictions on objects with standard structures. However, if the object contains some unusual structures, such as abnormal shape, tilt, or asymmetry, NPTC-net is able to generate smoother results that respect geometric information.

\subsection{Running Statistics}

\begin{table}[h!]
  \caption{Comparisons of number of parameters and FLOPs for classification.}
  \label{par}
  \centering
  \begin{tabular}{ccc}
    \hline
method  &  Parameters  &  FLOPs(Inference)\\
\hline
pointnet \cite{Qi_2017_CVPR}& 3.48M& 14.70B\\
pointnet++ \cite{NIPS2017_7095}&1.48M & 26.94B\\
pointcnn \cite{NIPS2018_7362}& \textbf{0.6M}& 25.30B\\
 Ours & 1.29M & \textbf{11.7B}\\
    \hline
  \end{tabular}
\end{table}

\begin{table}[h!]
  \caption{Comparisons of training time of networks on ModelNet40}
  \label{time}
  \centering
  \begin{tabular}{ccc}
\hline
Method & Settings & \makecell[c]{Accuracy and Training \\(+Pre-processing) time} \\
\hline
\multirow{2}{2.5cm}{PointNet++ \cite{NIPS2017_7095}} & adam, 1024 points & 90.7\%, 6 hours \\
                   & adam, 5000 points & 91.9\%, 20 hours \\
\hline
\multirow{2}{2.5cm}{ours} & adam, 2048 points & 92.4\%, 6h (+1.5h) \\
                   & SGD, 2048 points & 92.7\%, 12h (+1.5h) \\
\hline
  \end{tabular}
\end{table}

As shown in Table \ref{par}, we summarize our running statistics based with model for ModelNet40 with batch size 16. In comparison with several other methods, although we use ResNet structure, the fewer channels, smaller kernels and simpler interpolation (nearest neighboring) make NPTC use similiar parameters and even fewer FLOPs.

Total pre-processing time mainly depends on the grid's density. For most cases, the resolution of $100^3$ is enough to describe the shape of the point cloud, which is what we chose for our experiments. We also remark that the whole computation cost of constructing convolution is linearly depending on the resolution of voxel and the size of data. On a PC with Core i7-7700 CPU, the pre-processing takes about \textbf{0.5} seconds per point cloud with Matlab, and the whole dataset of ModelNet40 (12,311 shapes with 10,000 points each) takes only about 1.5 hours. It is negligible compared to the training of the deep neural networks and acceptable to do inference in practice.

\subsection{Feature Visualization}

\begin{figure}[h!]
  \centering

  \includegraphics[width=360pt]{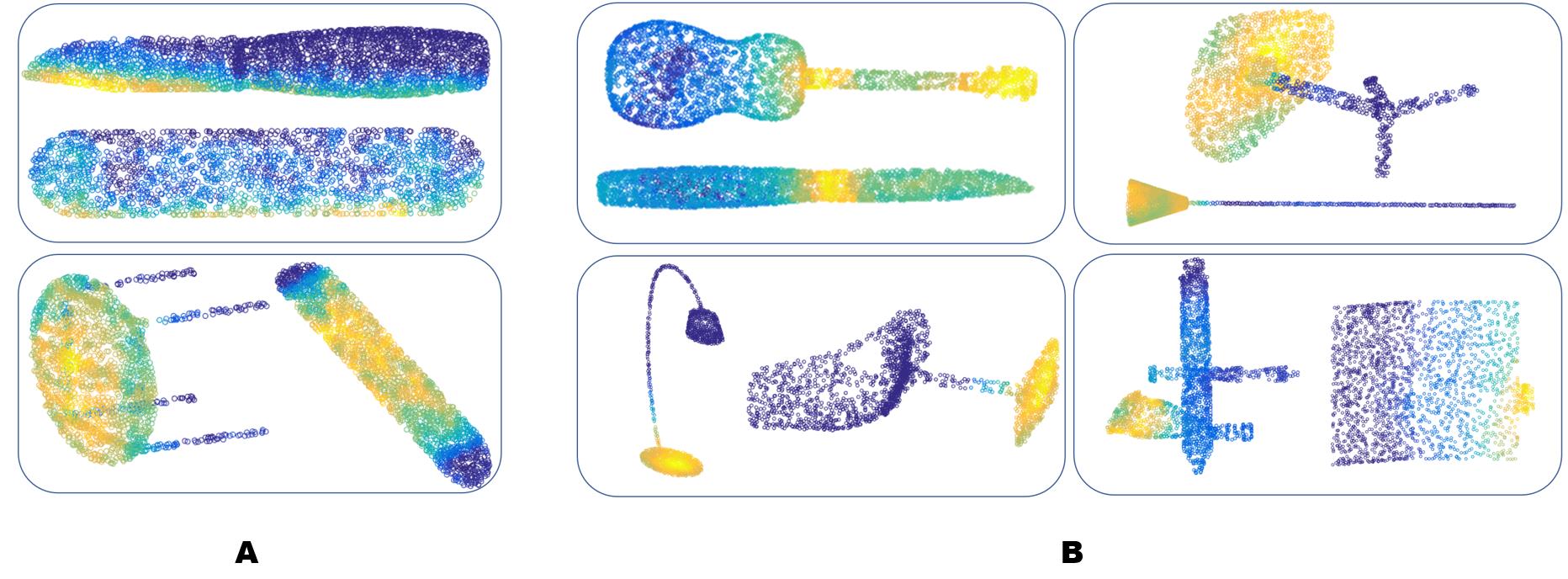}
  \caption{Feature Visualization: each feature from low (A) or high (B) level is displayed on 2 point clouds from different categories. High-level activations are in yellow and low-level activations are in blue.}
\label{vist}
\end{figure}

To visualize the effects of the proposed NPTC in the NPTC-net, we trained the network on ShapeNet Part and visualize learned features by coloring the points according to their level of activation. In Figure \ref{vist}, filters from the the first Convolution layer in the the first Residual block and final Convolution layer in the second Residual block are chosen. In order to easily compare the features at different levels, we interpolate them on the input point cloud. Observe that low-level features mostly represent simple structures like edges (top of (A)) and planes (bottom of (A)) with low variation in their magnitudes. In deeper layers, features are richer and more distinct from each other, like bottleneck (upper left of (B)), "big-head"(upper right of (B)), plane base (lower left of (B)), bulge (lower right of (B)).

\section{Conclusion}
\label{sec:conclusion}

This paper proposed a new way of defining convolution on point clouds, called the narrow-band parallel transport convolution (NPTC), based on a point cloud discretization of a manifold parallel transport. The parallel transport was defined specifically by a vector field generated by the gradient field of a distance function on a narrow-band approximation of the point cloud. The NPTC was used to design a convolutional neural network (NPTC-net) for point cloud classification and segmentation. Comparisons with state-of-the-art methods indicated that the proposed NPTC-net is competitive with the best existing methods.

\bibliographystyle{siamplain}
\bibliography{references}
\end{document}